\documentclass[11pt]{article}
\usepackage[utf8]{inputenc}
\usepackage[T1]{fontenc}
\usepackage{mathptmx}
\usepackage[a4paper,margin=1in]{geometry}
\usepackage{amsmath,amssymb}
\usepackage{graphicx}
\usepackage{booktabs}
\usepackage{array}
\usepackage{rotating}
\newcolumntype{C}[1]{>{\centering\arraybackslash}p{#1}}
\usepackage{microtype}
\usepackage{xcolor}
\usepackage[font=small,labelfont=bf,labelsep=period]{caption}
\usepackage[hidelinks]{hyperref}
\usepackage{url}
\usepackage{authblk}
\usepackage{setspace}

\newif\ifshowverify
\showverifyfalse

\title{\textbf{Faithful Explanations under Missing Modalities in Multimodal Dementia Prognosis}}
\author[1]{Firuz Kamalov\thanks{Corresponding author: \href{mailto:firuz@cud.ac.ae}{firuz@cud.ac.ae}}, for the Alzheimer's Disease Neuroimaging Initiative\thanks{Data used in preparation of this article were obtained from the Alzheimer's Disease Neuroimaging Initiative database, \url{adni.loni.usc.edu}. As such, the investigators within ADNI contributed to the design and implementation of ADNI and provided data but did not participate in the analysis or writing of this report. A complete listing of ADNI investigators can be found at \url{http://adni.loni.usc.edu/wp-content/uploads/how_to_apply/ADNI_Acknowledgement_List.pdf}.}}
\affil[1]{Canadian University Dubai, City Walk, P.O. Box 117781, Dubai, United Arab Emirates}
\date{}

\begin{document}
\maketitle

\begin{abstract}
\noindent Multimodal models for dementia prognosis are increasingly paired with large language models that translate feature attributions into clinician-facing narratives. In practice, costly modalities such as amyloid PET and cerebrospinal fluid biomarkers are frequently unavailable, and whether such explanations remain faithful under missing evidence has not been tested. We separate two failure points: phantom attribution, in which the fusion model assigns explanatory credit to modalities that were never observed, and unfaithful narration, in which the report omits missing modalities, cites them as evidence, or misstates whether acquiring them would change the prediction. Using 307 participants of the Alzheimer's Disease Neuroimaging Initiative with complete data, we masked every subset of four costly modalities, predicted 36-month conversion from mild cognitive impairment to Alzheimer's disease, and restored the masked modalities to obtain interventional ground truth. Across five missing-data strategies, phantom attribution accounted for 12\% to 23\% of total attribution mass and rose to 28\% to 45\% when four modalities were absent; no strategy eliminated it, and a missingness indicator was lowest. Across 51,200 narratives from four language models under four prompting regimes, unconstrained prompting acknowledged absent modalities in 48\% of cases for open-weight narrators; naming the absent modalities in the prompt raised acknowledgement to 88\% and reduced phantom evidence twenty-fold, but statements about the value of acquiring a missing test were correct in only 27\% to 41\% of cases until a claim-verification loop raised accuracy to 69\% for open-weight and 99.9\% for a proprietary narrator. Each unit of upstream phantom attribution doubled the odds of phantom evidence in the narrative. The rule-based verifier missed most direction errors; re-measured by a language-model judge calibrated against clinician-validated annotations, the attribution fidelity of open-weight narrators was 0.86 to 0.88 under every regime, verification did not raise it, and it rose with narrator scale and fell with upstream phantom attribution. Annotation also revealed a failure outside the taxonomy, declaring an observed modality absent, which occurred in 3.0\% of complete-data reports under prose prompting and never under a structured template. Results replicated on naturally incomplete records. We release the masking protocol and evaluation code.
\end{abstract}

\noindent\textbf{Keywords:} multimodal medical AI; missing modalities; explainable AI; large language models; explanation faithfulness; clinical decision support; Alzheimer's disease; mild cognitive impairment

\section{Introduction}
\label{sec:1}

The prognosis of mild cognitive impairment (MCI) is a multimodal problem. Contemporary biological definitions of Alzheimer's disease (AD) rest on amyloid, tau, and neurodegeneration markers that are measured by positron emission tomography (PET), cerebrospinal fluid (CSF) assays, and structural magnetic resonance imaging (MRI), and the risk of progression to dementia is further modulated by genotype, cognition, and demographics \cite{ref1}. Research cohorts such as the Alzheimer's Disease Neuroimaging Initiative (ADNI) were assembled precisely so that these modalities could be studied together \cite{ref2}. Machine learning models trained on such cohorts now reach clinically interesting accuracy for predicting conversion from MCI to AD, and the natural next step is to place them in front of clinicians.

Two developments shape how that step is being taken. The first is the demand for explanation. High-stakes deployment has been argued to require models whose decisions can be inspected, and the limits of post hoc explanation in medicine have been documented with equal force \cite{ref3,ref4}. In practice, most deployed systems compromise on feature attribution, most often Shapley-value methods, which decompose a prediction into additive contributions of its inputs \cite{ref5}. The second development is the use of large language models (LLMs) to translate those attributions into prose. Several recent systems accept a table of attributions and return a narrative that a clinician can read in seconds, and early evaluations report that such narratives are preferred to raw attribution plots but are prone to fabricated or misstated claims \cite{ref6,ref7}.

This paper concerns a condition that is ubiquitous in practice and almost absent from the evaluation of these systems: the patient in front of the clinician is missing some of the modalities the model was trained on. Amyloid PET is expensive and unevenly available, lumbar puncture is invasive and frequently declined, and genotyping is not routine. A prognostic model intended for use will therefore be asked to predict, and to explain, from incomplete evidence. The statistical literature has long distinguished missing data handled for description from missing data handled for prediction, and has shown that imputation strategies that are principled for the former can be inappropriate for the latter \cite{ref8,ref9}. What has not been examined is what missingness does to the explanation.

We argue that a missing modality can corrupt an explanation at two separate points, and that the two must be measured separately. The first point is the attribution itself. When a missing modality is filled by a mean value, a multiply imputed draw, or a learned placeholder, the substituted input is still an input, and an attribution method will in general assign it credit. We call this phantom attribution: explanatory weight placed on evidence that was never observed. Phantom attribution is not a property of the language model. It is a property of the fusion architecture and of the missing-data strategy, and it occurs before any narrative is written. The second point is the narrative. Even when the attribution is clean, a language model asked to narrate it may fail to mention that a modality was absent, may cite the absent modality as evidence, may express confidence that is unwarranted given what was observed, or may state that acquiring the missing test would or would not change the assessment when the opposite is true. The last of these failures is specific to the clinical setting, where the most useful sentence an explanation can contain is often a statement about which test to order next.

Evaluating either failure requires ground truth about what the absent evidence would have contributed. We obtain it by intervention. Within the subset of ADNI participants for whom every modality is observed, we mask each possible subset of modalities, predict and explain from the remainder, and then restore the masked modalities and observe what actually changes. This yields, for every masked patient and every narrative claim about a missing modality, the executed counterpart of the claim. The design follows the logic of interventional faithfulness evaluation, in which an explanation's account of what a model depends on is compared with the model's behaviour under the matching intervention rather than with human intuition \cite{ref10}. A recent study applied this logic to vision-language models on general benchmarks and found that self-explanations systematically overstate the sufficiency of the evidence that remains when a modality is removed \cite{ref11}. We extend the logic to a clinical prognosis task with tabular, imaging-derived, and biomarker modalities, to attributions produced by a separate explanation method rather than by the language model itself, and to statements about the value of acquiring what is missing.

The contributions of this work are as follows. First, we formalise phantom attribution and narrative faithfulness under missing modalities as measurable quantities with interventional ground truth. Second, we compare five missing-data strategies for a multimodal MCI-to-AD prognostic model by the attribution mass they assign to absent modalities, holding the attribution method fixed, and show that no strategy eliminates it and that the quantity is sensitive to the attribution reference in a way that the pre-specified ablation exposes. Third, we evaluate three open-weight language models and one proprietary model under four narration regimes of increasing constraint on five faithfulness metrics across 51,200 reports, and show that the regime matters more than the model for acknowledgement and phantom evidence, that neither matters enough for statements about which test to order next until the claims are verified against the model, that upstream phantom attribution propagates into the narrative, lowering the fidelity of direction claims as well as inviting phantom evidence, that verification against the model corrects acquisition claims but not direction claims, which a calibrated language-model judge shows to depend on the scale of the narrator, and that a prompt which asks the narrator to name absent modalities leads small narrators to name some when none are absent. Fourth, we release the masking protocol, the claim-extraction and verification code, and the evaluation harness so that the study can be reproduced on any cohort with a data use agreement, and so that reporting of missing-data handling in explainable prognostic models can move toward the standards already expected for the predictions themselves \cite{ref12}.

The remainder of the paper is organised as follows. Section~\ref{sec:2} reviews related work. Section~\ref{sec:3} formalises the problem. Section~\ref{sec:4} describes the cohort, the models, the narration regimes, the metrics, and the statistical analysis, and records every deviation from the pre-specified protocol. Section~\ref{sec:5} reports the results. Sections~\ref{sec:6} to~\ref{sec:8} discuss the implications, the limitations of the design, and the conclusions.

\section{Related Work}
\label{sec:2}

\subsection{Explainable Multimodal Prognosis in Alzheimer's Disease}
\label{sec:2.1}

Prediction of conversion from MCI to AD from combined imaging, fluid, genetic, and cognitive data is a mature research problem, and explainability has become a standard component of it. Tree-based models with Shapley attributions have been used to expose pairwise interactions between modalities in conversion prediction \cite{ref13}, and hierarchical designs that attribute first across modalities and then within them have been proposed to make the resulting explanations easier to read \cite{ref14}. A structured review of counterfactual, longitudinal, and multimodal explanation methods for MRI-based diagnosis found that external validation is uncommon, that calibration is inconsistently reported, and that structured reporting with explicit provenance for each numeric claim is rare \cite{ref15}. None of the studies reviewed evaluates explanations when a modality is absent at inference time, and the review does not list missingness among the axes on which methods are compared. Related work has combined plasma biomarkers with machine learning to model clinical progression in cognitively normal individuals at risk for AD \cite{ref16}, compared volumetric features with three-dimensional image tensors across cohorts for AD classification \cite{ref17}, and proposed explainable AI informed by the Research Domain Criteria as a paradigm for multilevel AD diagnosis and progression prediction \cite{ref18}. Our work addresses that gap directly.

\subsection{Learning and Predicting with Missing Modalities}
\label{sec:2.2}

Missing modalities have been studied extensively as a robustness problem. Early work trained a Bayesian meta-learning model to reconstruct a missing modality's latent representation from the modalities present \cite{ref19}. Later designs separate modality-shared from modality-specific features so that a missing modality can be substituted by a shared representation at inference \cite{ref20}, and healthcare-specific architectures such as HEALNet keep a per-modality encoder and a shared latent space so that any subset of modalities can be fused without imputation \cite{ref21}. These methods are evaluated by predictive performance under increasing rates of missingness. Whether the resulting models attribute their predictions to the modalities actually present is not examined.

The statistical treatment of missing data supplies the background for the strategies we compare. Rubin's taxonomy distinguishes data missing completely at random, at random, and not at random \cite{ref22}, and multiple imputation by chained equations is the standard method for the second case \cite{ref23}. For prediction rather than estimation, Sperrin and colleagues argue that the goal is a model that performs well on the incomplete data it will actually see, which favours training with the same missingness the model will encounter and, in some settings, adding missingness indicators \cite{ref8}. Josse and colleagues show that simple imputation followed by a sufficiently flexible learner is consistent for prediction, which explains why mean imputation remains widespread \cite{ref9}. The five strategies we compare span this range.

\subsection{Attribution in Multimodal Models}
\label{sec:2.3}

Shapley values provide an axiomatic decomposition of a prediction into contributions of its inputs \cite{ref5}, and grouping inputs by modality yields a modality-level attribution. Gapp and colleagues proposed an occlusion-based, model-agnostic measure of modality contribution and applied it to image-text and image-tabular clinical tasks, but studied only patients with complete data \cite{ref24}. MultiSHAP extends Shapley analysis to cross-modal interactions \cite{ref25}. A separate line of work has established that attribution methods can produce plausible-looking maps that are insensitive to the model or the data \cite{ref26}, and has proposed removing attributed features and retraining as a check on whether attributions identify what the model uses \cite{ref27}. Phantom attribution as defined here is a specific instance of the same concern: attribution assigned to an input that carries no observed information.

\subsection{Language Models as Explanation Narrators}
\label{sec:2.4}

The practice of feeding attribution tables to a language model and asking for prose is recent and spreading. A prompt framework for narrating machine learning explanations in intensive care reported that clinicians preferred the narratives but that the narratives contained unsupported statements \cite{ref7}. Ichmoukhamedov and colleagues proposed the first automated metrics for such narratives and found that hallucination of feature effects was common across models and prompts \cite{ref6}. In dementia specifically, multi-agent language-model pipelines have been applied to longitudinal clinical notes \cite{ref28}, and tabular language models have been used for few-shot prediction with self-generated rationales \cite{ref29}. Clinical summarisation studies have developed frameworks for classifying hallucinations and omissions by clinical severity \cite{ref30}. Missing evidence is not a category in any of these frameworks.

\subsection{Faithfulness and Its Evaluation}
\label{sec:2.5}

Jacovi and Goldberg distinguish plausibility, the degree to which an explanation convinces a reader, from faithfulness, the degree to which it reflects the reasoning of the model, and argue that only the latter can be evaluated without human judgement \cite{ref10}. For explanations generated by language models, faithfulness has been probed by perturbing inputs and checking whether the stated reasoning tracks the change in the output \cite{ref31,ref32}. The claim-level decomposition we adopt follows the approach used to score long-form factual generation, in which a text is split into atomic claims and each claim is verified against a source \cite{ref33}. The closest work to ours is the interventional protocol of Javadov and colleagues, which compares a vision-language model's account of its dependence on a modality with its behaviour when that modality is removed or restored \cite{ref11}. Their study is confined to general-domain vision-language benchmarks, to self-explanation by the same model that makes the prediction, and to modality-level claims. We study a clinical prognosis task, a separate attribution method narrated by a language model, tabular and biomarker modalities, and claims about the value of acquiring what is missing.

\section{Problem Formulation}
\label{sec:3}

\subsection{Notation}
\label{sec:3.1}

Let a patient be described by K modalities, and let the observed data be $x = (x_1, \dots, x_K)$, where $x_k$ is the feature vector of modality k. A missingness pattern is a subset M of $\{1, \dots, K\}$ indexing the modalities that are absent, and O denotes its complement, the modalities that are observed. A missing-data strategy s maps the observed modalities and the pattern to a complete input, written $x^{(s)}_M$, in which each absent modality has been replaced by a substitute that depends on the strategy. The prognostic model f maps a complete input to a probability of conversion within 36 months. We write $p_M = f(x^{(s)}_M)$ for the prediction under pattern M and $p_\emptyset$ for the prediction from complete data.

An attribution method assigns to each modality k a contribution $\varphi_k(x^{(s)}_M)$ such that the contributions sum to the difference between the prediction and a reference value. We use modality-level Shapley values, with modality k treated as a single player and absent from a coalition by replacement with the strategy's substitute, so that the attributions satisfy

\begin{equation}
p_M - p_{\mathrm{ref}} = \sum_{k=1}^{K} \varphi_k\!\left(x^{(s)}_M\right).
\end{equation}

A narrator g maps the attribution vector, the prediction, and the list of observed and absent modalities to a text T. A claim extractor decomposes T into atomic claims $c_1, \dots, c_n$, each tagged with the modalities it mentions and with a claim type drawn from the set {evidence, direction, ranking, confidence, acquisition}.

\subsection{Phantom Attribution}
\label{sec:3.2}

An absent modality carries no information about the patient, so a faithful attribution assigns it none. In practice the substitute for an absent modality is still an input to f, and any attribution that treats the substitute as an ordinary feature will in general give it non-zero credit. We define the phantom attribution mass of a strategy s on a patient with pattern M as the share of total absolute attribution assigned to absent modalities,

\begin{equation}
\mathrm{PAM}_s(x, M) = \frac{\sum_{k \in M} \left|\varphi_k\!\left(x^{(s)}_M\right)\right|}{\sum_{k=1}^{K} \left|\varphi_k\!\left(x^{(s)}_M\right)\right|}.
\end{equation}

A strategy is free of phantom attribution when the mass is zero for every patient and every pattern. This quantity is a property of the pair (f, s) and of the attribution method, and it is measured before any narrative is produced. Note that a strategy can have zero phantom attribution and still be a poor predictor, and that a strategy with substantial phantom attribution can still predict well, since the substitute may carry population information that improves $p_M$. The two properties must be reported together.

\subsection{Narrative Faithfulness under Missingness}
\label{sec:3.3}

We distinguish four ways in which a narrative can misrepresent what was observed.

The first is omission. A faithful narrative names each absent modality as absent. Missingness acknowledgement recall is the fraction of modalities in M that the narrative identifies as unavailable.

The second is phantom evidence. A narrative cites phantom evidence when a claim of type evidence, direction, or ranking names a modality in M as though it had been observed. The phantom-evidence rate is the fraction of such claims among all claims that name a modality.

The third is attribution infidelity. Among claims that name an observed modality, a claim is faithful if its stated direction and relative importance agree with the attribution vector. Attribution fidelity is the fraction of faithful claims.

The fourth is a misstatement about the value of acquiring what is missing. A claim of type acquisition asserts that obtaining modality k in M would, or would not, materially change the assessment. Because the masked patients have complete data, the counterpart of this claim can be executed: we restore $x_k$, recompute the prediction, and observe whether the change exceeds a pre-specified threshold. The value-of-information accuracy is the agreement between stated and executed change across all acquisition claims. Formally, for a claim that restoring modality k would change the prediction, with $\delta$ a clinically motivated threshold on the probability scale, the claim is correct when

\begin{equation}
\left| f\!\left(x^{(s)}_{M \setminus \{k\}}\right) - f\!\left(x^{(s)}_M\right) \right| \geq \delta ,
\end{equation}

and a claim that restoring k would not change the prediction is correct when the inequality is reversed.

\subsection{Confidence Calibration of Narratives}
\label{sec:3.4}

A narrative typically conveys a level of confidence in words. As modalities are removed, the epistemic uncertainty of the model should rise and the stated confidence should fall. We map the narrative's confidence language to an ordinal scale and compare it with a model-side uncertainty measure computed from an ensemble of f, so that narratives which remain confident as evidence disappears can be identified. This quantity is reported separately because it depends on the model's uncertainty estimates as well as on the narrator.

\subsection{Interventional Ground Truth}
\label{sec:3.5}

Every quantity above except omission is defined with reference to either the attribution vector or the model's behaviour under a matching intervention, not to a human judgement of plausibility. This is the sense of faithfulness used throughout \cite{ref10}. The masking protocol supplies the interventions. For each participant with complete data and for each non-empty pattern M, the model predicts and the narrator explains from $x^{(s)}_M$, and the executed effects of restoring each k in M are computed from the same participant's complete record. The human annotation described in Section~\ref{sec:4} is used to validate the automatic claim extractor, not to define the metrics.

\section{Materials and Methods}
\label{sec:4}

\subsection{Cohort and Outcome}
\label{sec:4.1}

Data used in the preparation of this article were obtained from the Alzheimer's Disease Neuroimaging Initiative database, adni.loni.usc.edu. ADNI was launched in 2003 as a public-private partnership led by Principal Investigator Michael W. Weiner, MD. Its primary goal has been to test whether serial magnetic resonance imaging, positron emission tomography, other biological markers, and clinical and neuropsychological assessment can be combined to measure the progression of mild cognitive impairment and early Alzheimer's disease \cite{ref2}. Participants are included if they carry a diagnosis of MCI at baseline and have at least one diagnostic visit at or beyond 36 months, or convert to dementia before that point. The outcome is conversion to a clinical diagnosis of AD dementia within 36 months of baseline. Participants who revert to normal cognition or who are lost to follow-up before 36 months without converting are excluded from the primary analysis and retained for a sensitivity analysis in which they are treated as non-converters.

Six modalities are used. Structural MRI is represented by FreeSurfer-derived cortical thickness and subcortical volumes normalised by intracranial volume. Amyloid PET is represented by the centiloid value and its positivity indicator. CSF comprises A$\beta$42, phosphorylated tau, total tau, and their ratios. Genotype is the APOE $\varepsilon$4 allele count. Cognition comprises the Mini-Mental State Examination, the Alzheimer's Disease Assessment Scale cognitive subscale, the Clinical Dementia Rating sum of boxes, and the Rey Auditory Verbal Learning Test. Demographics comprise age, sex, and years of education. Demographics and cognition are treated as always observed, since they are available in any clinical encounter, so that masking is applied to the four costly modalities MRI, PET, CSF, and genotype, giving fifteen non-empty missingness patterns.

Two analytic cohorts are defined. The complete-data cohort comprises participants for whom all six modalities are observed at baseline; it supplies the interventional ground truth. The naturally incomplete cohort comprises all eligible participants regardless of which modalities are observed; it supplies a test under the missingness that ADNI actually exhibits, which is not at random because PET and CSF acquisition depend on study phase, site, and participant consent.

\begin{figure}[htbp]
\centering
\includegraphics[width=0.95\textwidth]{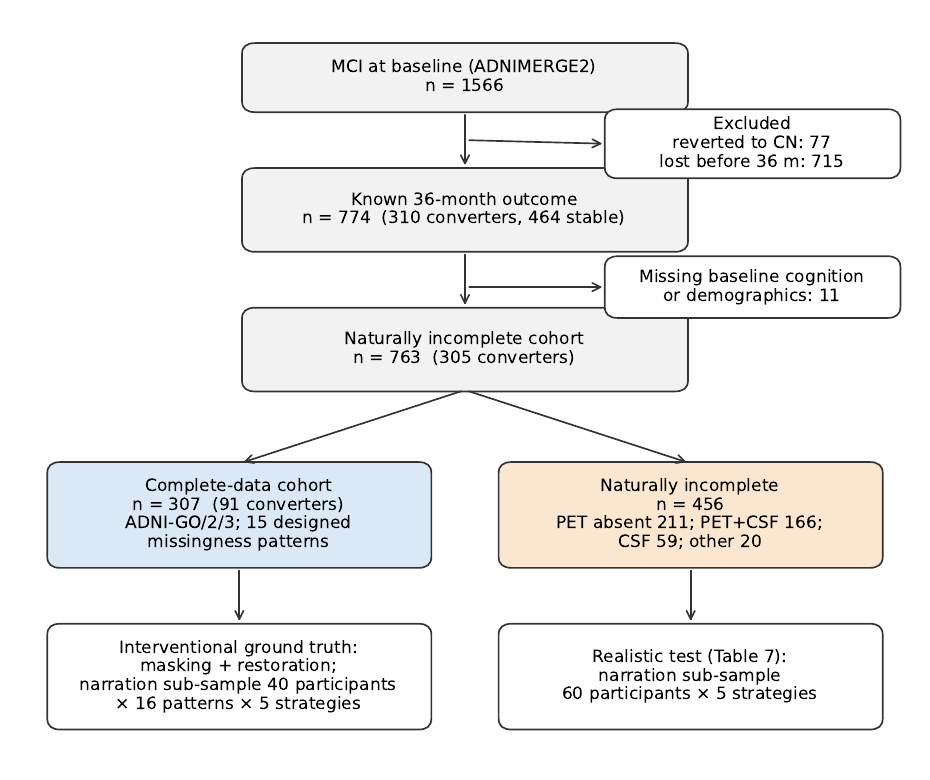}
\caption{Participant flow and analytic cohorts. The complete-data cohort supplies the interventional ground truth through designed masking and restoration; the naturally incomplete cohort supplies a test under the missingness that ADNI exhibits.}
\label{fig:1}
\end{figure}

Figure~\ref{fig:1} gives the flow. Of 1,566 participants with MCI at baseline in the ADNIMERGE2 tables, 715 were lost to follow-up before 36 months without converting and 77 reverted to normal cognition, leaving 774 with a known 36-month outcome, of whom 310 converted. Eleven lacked baseline cognition or demographics, giving a naturally incomplete cohort of 763 participants with 305 converters. Among them PET was absent for 382, CSF for 232, MRI for 20, and genotype for 1; the most common patterns were PET alone in 211 participants, PET with CSF in 166, and CSF alone in 59. The complete-data cohort comprises the 307 participants with every modality observed, of whom 91 converted; all were enrolled in ADNI-GO, ADNI-2, or ADNI-3, all had 3T MRI, 268 had florbetapir and 39 florbetaben PET, their mean age was 71.8 years with standard deviation 7.2, and 41\% were women. The 456 participants with at least one absent modality form the naturally incomplete test set.

\subsection{Prognostic Model}
\label{sec:4.2}

The prognostic model is a modality-tokenised transformer. Each modality is encoded by a small multilayer perceptron into a token of fixed dimension, the tokens are concatenated with a learned classification token and a learned modality-identity embedding, and a two-layer transformer encoder with multi-head self-attention produces the classification token that is read out by a linear layer \cite{ref34}. The architecture is deliberately simple so that differences between missing-data strategies are not confounded by architectural capacity. The same architecture is used for every strategy; only the treatment of absent modalities differs.

An ensemble of five models with different initialisations and data orderings is trained for each strategy. The ensemble mean is the prediction and the ensemble standard deviation is the uncertainty measure used in Section~\ref{sec:3.4}. Probabilities are calibrated by temperature scaling on the validation fold \cite{ref35}.

\subsection{Missing-Data Strategies}
\label{sec:4.3}

Five strategies are compared. Under mean imputation, each feature of an absent modality is replaced by its training-set mean. Under multiple imputation, twenty draws are generated by chained equations conditioned on the observed modalities \cite{ref23}, the model predicts on each, and the predictions and attributions are averaged. Under the placeholder-token strategy, an absent modality's token is replaced by a learned vector specific to that modality, and the model is trained with random modality dropout so that the placeholder is seen during training. Under the dropout-only strategy, the absent modality's token is removed from the sequence entirely, so that the transformer attends over a shorter sequence, and the model is again trained with random modality dropout. Under the indicator strategy, mean imputation is combined with a binary missingness indicator appended to each modality token, following the recommendation for prediction models that will encounter missingness at deployment \cite{ref8}.

The strategies differ in whether the substitute for an absent modality is a fixed feature vector, a stochastic draw, a learned vector, no vector at all, or a fixed vector accompanied by a flag. This ordering corresponds to a decreasing opportunity for the attribution method to assign credit to the substitute, and the first research question tests whether phantom attribution follows the ordering.

\subsection{Attribution}
\label{sec:4.4}

Modality-level Shapley values are computed with modalities as players \cite{ref5,ref36}. With six modalities, exact computation over all coalitions is feasible. A modality absent from a coalition is represented by the strategy's own substitute, so that the attribution is measured on the same input the model actually consumed. Within-modality feature attributions are computed by the same method with features as players, conditional on the modality-level result, and are used only for the attribution fidelity metric. All attributions are computed on the ensemble mean prediction. Two reference values are used and reported side by side. The pre-specified reference is the prediction for an input in which every modality takes its training-set mean. The second, introduced by the pre-specified ablation, is the expectation of the prediction over a background sample of training participants, so that a coalition's missing players are drawn from the data rather than fixed at the mean. The two references give different answers for the mean-imputation strategy for a reason set out in Section~\ref{sec:5.1}, and the second is used for the figures.

\subsection{Narration Regimes}
\label{sec:4.5}

A narrator receives a structured input comprising the predicted probability, the ensemble uncertainty, the modality-level attribution vector with signs, the top three within-modality features per observed modality, and the list of observed modalities. It is asked to produce a report of at most 200 words for a memory clinic physician. Four regimes of increasing constraint are compared, and the regimes differ in whether and how the absence of a modality is communicated.

Under the unconstrained regime, the narrator receives the attribution table, in which an absent modality appears with whatever attribution the strategy produced for its substitute, and the list of observed modalities. Nothing states that a modality is absent; this simulates a pipeline that computes attributions on the completed feature matrix and does not carry missingness metadata to the narrator. Under the explicit-missingness regime, the prompt additionally names the absent modalities in a separate sentence and instructs the narrator to say so, which places a possible conflict between the flag and the attribution table in front of the narrator. Under the structured regime, the narrator must fill a template with fixed fields for the assessment, the supporting evidence per observed modality, the modalities not available, and the expected effect of obtaining each. Under the verified regime, the structured output is passed to the claim extractor and verifier of Section~\ref{sec:4.6}; any claim that fails verification is returned to the narrator with the reason, and the narrator is asked to revise, for up to three rounds.

Four narrators are evaluated. Three are open-weight instruction-tuned models of the Qwen2.5 family at 3, 7, and 14 billion parameters, run locally with greedy decoding for the primary analysis \cite{ref37}. The fourth is Claude Sonnet 5.5, a proprietary model accessed through its provider's API at default sampling settings \cite{ref38}; it was added after the open-weight results showed a strong dependence on model scale, and its results are reported separately. The structured input given to every narrator contains, in addition to the prediction and the attributions, the values of up to three participant-level measurements per observed modality. The proprietary model was accessed under a zero-data-retention agreement with the provider, under which inputs are neither stored nor used for training, which meets the conditions that the ADNI data use agreement places on processing by third-party services.

The narration study uses a stratified sub-sample of 40 participants from the complete-data cohort, 12 of them converters, drawn from the held-out folds of the first cross-validation repetition. Each participant is narrated under all 16 missingness patterns including the complete pattern, all five strategies, and all four regimes, giving 3,200 reports per narrator and 12,800 across the four regimes; the three open-weight narrators contribute 38,400 reports and the proprietary narrator 12,800, for 51,200 in total. A further 60 participants from the naturally incomplete cohort are narrated under their actual missingness pattern, all five strategies, and all four regimes, giving 1,200 reports per narrator.

\subsection{Claim Extraction and Verification}
\label{sec:4.6}

Each report is decomposed into atomic claims by a separate language model prompted with the claim taxonomy of Section~\ref{sec:3.1} and with examples, following the decomposition approach used in long-form factuality evaluation \cite{ref33}. Each claim is tagged with the modalities it names and its type. A rule-based verifier then checks each claim against the structured input and, for acquisition claims, against the executed restoration. An evidence, direction, or ranking claim that names an absent modality is flagged as phantom evidence. A direction claim about an observed modality is verified against the sign of its attribution; a ranking claim against the order of absolute attributions. An acquisition claim is verified against the executed change in prediction with threshold $\delta$ = 0.10 on the probability scale; sensitivity to $\delta$ in {0.05, 0.10, 0.15} is reported. A confidence claim is mapped to one of four ordinal levels by a fixed lexicon and compared with the quartile of the ensemble uncertainty. Acknowledgement of an absent modality is detected by the claim extractor and, as a check, by a regular expression over the report text; the claim-based measure is primary. Reports that the extractor could not parse into claims are retained in the denominator for acknowledgement and excluded from claim-level metrics; the parse failure rate is reported. A sixth failure, identified during annotation and absent from the pre-specified taxonomy, is false absence: a report states that a modality was unavailable although it was observed. It is detected over the full corpus by a rule-based detector that finds statements of data unavailability, isolates the noun phrase they govern, and tests whether it names an observed modality; statements that a clinical finding is absent, such as the absence of amyloid positivity, are excluded. The detector was tuned on the 50 annotated reports, where it recovers all four false-absence reports with no false positives, and a random sample of 30 of its positives over the corpus was reviewed by hand and found to be correct in every case; its recall over the corpus is not known and the rates it yields are lower bounds.

\subsection{Validation of the Automatic Pipeline}
\label{sec:4.7}

The pre-specified protocol called for two human annotators to decompose 200 reports and for two memory clinic physicians to rate 50 reports for usefulness. This was not carried out before the present analysis, and the pipeline was instead validated against two independent language-model judges, following the practice of using a strong model as a reference annotator \cite{ref39}. The first judge is Qwen2.5-14B-Instruct, applied to a stratified sample of 180 reports from the open-weight narrators, 45 per regime and 60 per narrator. The second is Claude Sonnet 5.5, applied to 240 reports, 60 per regime and 60 per narrator including its own narrations. Each judge receives the report and the structured input and returns, independently of the pipeline, whether each absent modality was acknowledged, whether any absent modality was cited as evidence, whether any direction claim contradicted the attribution, the stated change for each acquisition claim, and the confidence level. Agreement between each judge and the pipeline, and between the two judges on the 180 reports they share, is reported as raw agreement and Cohen's $\kappa$ \cite{ref40}. In addition, a stratified sample of 50 reports from the open-weight narrators, 45 from the explicit-missingness regime and 5 from the structured regime, was annotated by Claude Fable 5.1, reading each report against its structured input under a written protocol that is released with the code, and recording, for each absent modality, whether its absence was stated; any passage citing an absent modality as evidence; any statement about the value of acquiring a missing modality; every claim whose stated direction of effect contradicts the signed attribution, including factual misstatements about the patient; the confidence wording; and a provisional usefulness rating from 1 to 5. A memory clinic physician then reviewed every annotation against the report and the structured input and confirmed all 50 without change. The validated sheet is released with the code and is the reference against which Section~\ref{sec:5.5} reports the pipeline and the judges.

Because the validation in Section~\ref{sec:5.5} showed that the rule-based verifier detects few direction errors, attribution fidelity was re-measured by a language-model judge. From each of the sixteen narrator-by-regime cells of the primary corpus, 200 reports were drawn at random with a fixed seed, and the 3,200 sampled reports were shuffled together with the 50 validated reports, which served as hidden calibration items, and divided into batches of 25. Batches were judged in random order until approximately 100 reports per cell had been judged, giving 1,626 sampled reports, 1,245 from the open-weight narrators and 381 from the proprietary narrator, and 24 calibration reports. The judge received the structured input that the narrator had received and the report, and returned, under the written protocol used for the validated annotation, the number of claims that state, or imply through a heading or list framing, the direction in which a named feature or modality moves the predicted risk; every such claim whose direction contradicts the signed attribution, and every factual misstatement about the patient that the input contradicts; and false absence, phantom evidence, acknowledgement, and acquisition statements. Claude Fable 5.1 judged the first 875 reports and Claude Opus 5.5 the remaining 775 under the same protocol, and Claude Opus 5.5 re-judged 100 reports already judged by Claude Fable 5.1, without access to the earlier judgements, to measure the effect of the change of judge. Judge-measured attribution fidelity is one minus the ratio of contradicted claims to direction claims, pooled within a cell, with 95\% percentile bootstrap intervals clustered by participant, and is modelled at the claim level with the hierarchical logistic model of Section~\ref{sec:4.8} with a random intercept for participant. Both judges were accessed under the zero-data-retention agreement described in Section~\ref{sec:4.5}, and the protocol and the identifiers of the sampled reports are released with the code.

\subsection{Experimental Design and Statistical Analysis}
\label{sec:4.8}

The complete-data cohort is split by participant into five folds. Models are trained on four folds and evaluated on the fifth, and the split is repeated with three different seeds, giving fifteen evaluation runs. Every prognostic metric is computed per run and summarised as the mean with a 95\% confidence interval from the percentile bootstrap over participants within runs and over runs. Phantom attribution mass is compared across strategies with a Friedman test over the fifteen runs followed by pairwise Wilcoxon signed-rank tests with Holm correction. Predictive performance is reported as the area under the receiver operating characteristic curve, the area under the precision-recall curve, the Brier score, and calibration slope, so that the cost of each missing-data strategy in prediction can be set beside its phantom attribution.

Narrative metrics are summarised as proportions with 95\% percentile bootstrap intervals over reports, clustered by participant. The pre-specified claim-level mixed-effects logistic regression was fitted as a Bayesian hierarchical logistic model estimated by variational inference, with regime, narrator, number of absent modalities, and upstream phantom attribution mass as fixed effects and random intercepts for participant and run. The model was fitted with the variational Bayes estimator of the BinomialBayesMixedGLM class in statsmodels, which estimates mixed-effects logistic models with crossed random intercepts \cite{ref41}. Odds ratios are reported with 95\% credible intervals. The open-weight narrators are pooled in the primary narrative tables and the proprietary narrator is reported separately, since it was added after the protocol was fixed.

The naturally incomplete sub-sample is evaluated with the models trained on the complete-data folds, beside the complete-data reports restricted to the same seven missingness patterns, so that the difference between designed and natural missingness can be read directly. Because restoration is impossible for naturally missing modalities, only acknowledgement, phantom evidence, attribution fidelity, and confidence calibration are reported on this sub-sample.

Three sensitivity analyses were run as pre-specified: the attribution reference was changed, the threshold $\delta$ was varied, and the narration prompt was rephrased and the unconstrained and structured regimes re-run on a 21\% sub-sample of 665 reports per narrator. Sampling at temperature 0.7 across five seeds was run for the structured regime on 328 reports per seed and narrator. The pre-specified ablation in which the verifier's $\delta$ is deliberately misspecified was not run. Code for the masking protocol, the models, the claim extractor and verifier, and the analysis will be released under an open licence; ADNI data cannot be redistributed and must be obtained under the ADNI data use agreement.

\section{Results}
\label{sec:5}

\subsection{Phantom Attribution across Missing-Data Strategies}
\label{sec:5.1}

Table~\ref{tab:1} reports predictive performance and phantom attribution mass for each strategy on the complete-data cohort, averaged over the fifteen non-empty missingness patterns and the fifteen runs. All five strategies predicted conversion with similar discrimination under missingness, with AUROC between 0.881 and 0.897 against 0.907 to 0.916 on complete data, and with calibration slopes between 0.90 and 1.00. Mean imputation was the weakest predictor and the indicator strategy the strongest, but the differences were within the confidence intervals.

\begin{sidewaystable}[htbp]
\centering
\small
\setlength{\tabcolsep}{4pt}
\caption{Predictive performance and phantom attribution mass by missing-data strategy on the complete-data cohort under designed missingness. Mean and 95\% confidence interval over fifteen runs. AUROC on complete data was 0.909, 0.907, 0.916, 0.913, and 0.915 for the five strategies in the order listed. PAM denotes phantom attribution mass under the training-set-mean reference and under the background-sample reference.}
\label{tab:1}
\begin{tabular}{lC{3.1cm}C{3.1cm}C{3.1cm}C{3.1cm}C{3.1cm}C{3.1cm}}
\toprule
Strategy & AUROC & AUPRC & Brier & Calibration slope & PAM, mean reference & PAM, background reference \\
\midrule
Mean imputation & 0.881 [0.855, 0.904] & 0.720 [0.658, 0.777] & 0.132 [0.116, 0.148] & 0.96 [0.82, 1.15] & 0.000 [0.000, 0.000] & 0.178 [0.154, 0.202] \\
Multiple imputation & 0.892 [0.866, 0.915] & 0.742 [0.678, 0.803] & 0.123 [0.107, 0.140] & 0.99 [0.85, 1.16] & 0.207 [0.197, 0.217] & 0.226 [0.216, 0.236] \\
Indicator & 0.897 [0.872, 0.922] & 0.748 [0.687, 0.819] & 0.121 [0.104, 0.138] & 1.00 [0.84, 1.20] & 0.118 [0.105, 0.133] & 0.117 [0.107, 0.127] \\
Placeholder token & 0.893 [0.865, 0.918] & 0.747 [0.681, 0.810] & 0.124 [0.107, 0.142] & 0.94 [0.80, 1.10] & 0.197 [0.170, 0.221] & 0.135 [0.120, 0.150] \\
Dropout only & 0.895 [0.869, 0.920] & 0.758 [0.691, 0.824] & 0.123 [0.106, 0.142] & 0.90 [0.77, 1.07] & 0.190 [0.164, 0.216] & 0.131 [0.116, 0.147] \\
\bottomrule
\end{tabular}
\end{sidewaystable}

Phantom attribution was present under every strategy and under both references, with one exception that is an artefact. Under the pre-specified mean reference, the mean-imputation strategy had a phantom attribution mass of zero to numerical precision, because the substitute for an absent modality is identical to the reference value of that modality, so that the Shapley value of the absent modality is zero by construction rather than by any property of the model. Under the background reference, the same strategy had a phantom attribution mass of 0.178. The remaining four strategies were far less sensitive to the reference: multiple imputation 0.207 against 0.226, the indicator strategy 0.118 against 0.117, the placeholder token 0.197 against 0.135, and dropout 0.190 against 0.131. The pre-specified hypothesis that the placeholder-token and dropout strategies would assign little or no attribution to absent modalities was not supported. Under either reference, between an eighth and a quarter of the total attribution mass was assigned to modalities that were not observed.

The Friedman test over runs rejected equality of phantom attribution mass across strategies under both references, with $\chi^2$ = 45.6 and p = $3 \times 10^{-9}$ under the mean reference and $\chi^2$ = 38.4 and p = $9 \times 10^{-8}$ under the background reference. Under the background reference, after Holm correction, multiple imputation exceeded every other strategy with p = 0.0006, mean imputation exceeded the indicator, dropout, and placeholder strategies with p between 0.003 and 0.041, and the indicator, placeholder, and dropout strategies did not differ from one another with p between 0.12 and 0.72. Under the mean reference the indicator strategy was below multiple imputation, placeholder, and dropout with p between 0.0006 and 0.0012, and those three did not differ from one another.

\begin{figure}[htbp]
\centering
\includegraphics[width=0.72\textwidth]{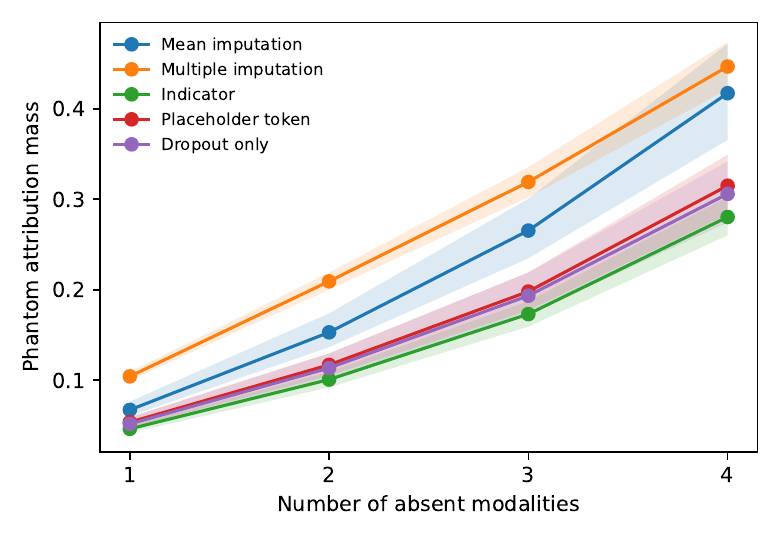}
\caption{Phantom attribution mass against the number of absent modalities, by missing-data strategy, under the background-sample reference. Bands are 95\% bootstrap intervals over the fifteen runs.}
\label{fig:2}
\end{figure}

\begin{figure}[htbp]
\centering
\includegraphics[width=0.72\textwidth]{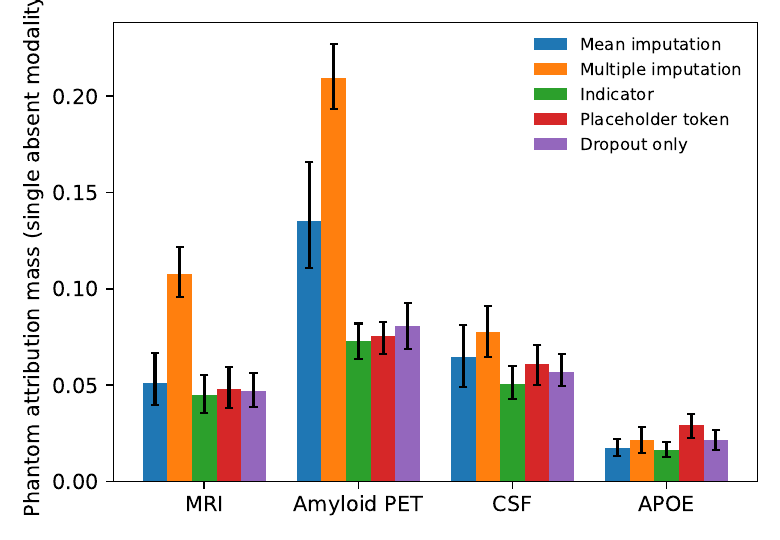}
\caption{Phantom attribution mass with one modality absent, by absent modality and missing-data strategy, under the background-sample reference. Error bars are 95\% bootstrap intervals.}
\label{fig:3}
\end{figure}

Phantom attribution mass rose approximately linearly with the number of absent modalities, as Figure~\ref{fig:2} shows for the background reference. With one modality absent it ranged from 0.046 for the indicator strategy to 0.104 for multiple imputation; with four absent it ranged from 0.280 to 0.447, and mean imputation reached 0.417. Figure~\ref{fig:3} breaks the one-modality case down by which modality is absent. Amyloid PET attracted the most phantom attribution under every strategy, from 0.073 for the indicator strategy to 0.209 for multiple imputation, followed by MRI and CSF at 0.045 to 0.108, with APOE genotype lowest at 0.016 to 0.029. The ordering matches the modalities' marginal effect on the prediction when restored, reported in Section~\ref{sec:5.3}.

\subsection{Narrative Faithfulness by Regime and Narrator}
\label{sec:5.2}

\begin{figure}[htbp]
\centering
\includegraphics[width=1.00\textwidth]{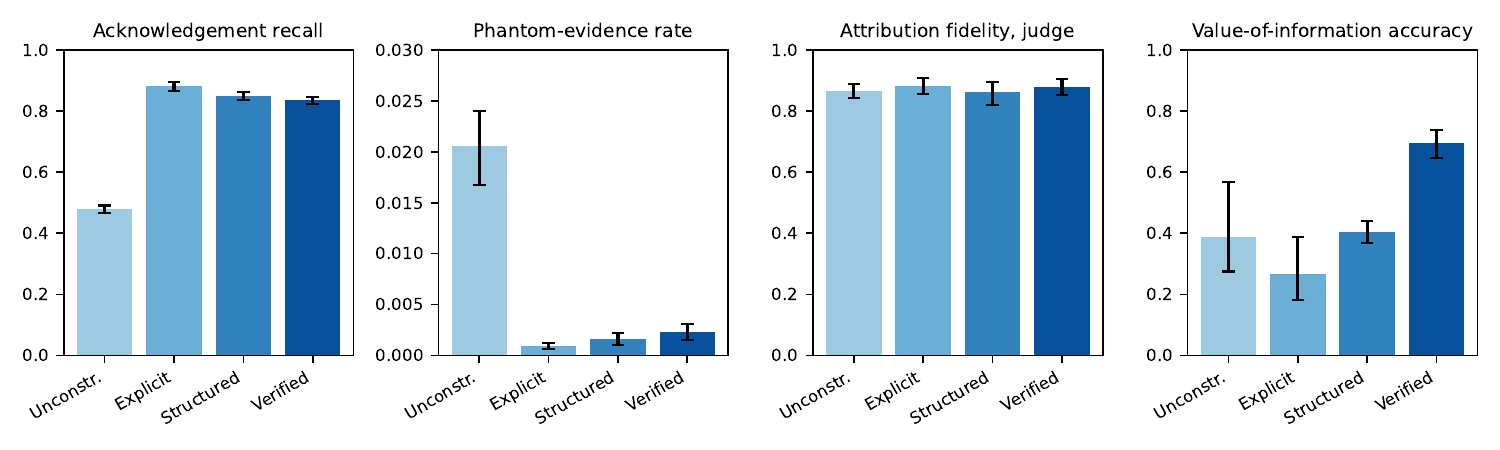}
\caption{Acknowledgement recall, phantom-evidence rate, attribution fidelity, and value-of-information accuracy by narration regime for the three open-weight narrators pooled, 9,600 reports per regime. Attribution fidelity is measured by the language-model judge on 308 to 320 sampled reports per regime. Error bars are 95\% bootstrap intervals clustered by participant.}
\label{fig:4}
\end{figure}

Table~\ref{tab:2} reports the narrative metrics for the three open-weight narrators pooled, 9,600 reports per regime, and separately for the proprietary narrator, 3,200 reports per regime. Figure~\ref{fig:4} plots the four principal metrics for the pooled open-weight narrators.

\begin{sidewaystable}[htbp]
\centering
\small
\setlength{\tabcolsep}{4pt}
\caption{Narrative faithfulness by regime. Rows 1 to 4 pool the three open-weight narrators, 9,600 reports per regime; rows 5 to 8 are the proprietary narrator, 3,200 reports per regime. Proportions with 95\% bootstrap intervals clustered by participant. Attribution fidelity is measured by the language-model judge of Section~\ref{sec:4.7} on a stratified random sample, with n the number of judged reports, as one minus the fraction of direction claims that contradict the signed attribution. For value-of-information accuracy, n is the number of acquisition claims on which it is computed. Confidence calibration is Spearman's correlation between stated confidence level and ensemble-uncertainty quartile; a negative value is the desired direction.}
\label{tab:2}
\begin{tabular}{p{3.0cm}C{3.8cm}C{3.8cm}C{3.8cm}C{3.8cm}C{3.8cm}}
\toprule
Regime & Acknowledgement recall & Phantom-evidence rate & Attribution fidelity, judge & Value-of-information accuracy & Confidence calibration \\
\midrule
Unconstrained & 0.479 [0.466, 0.491] & 0.0206 [0.0167, 0.0240] & 0.866 [0.844, 0.888], n = 308 & 0.387 [0.275, 0.569], n = 217 & $-$0.43 [$-$0.58, $-$0.32] \\
Explicit missingness & 0.881 [0.866, 0.896] & 0.0009 [0.0006, 0.0012] & 0.882 [0.856, 0.907], n = 308 & 0.268 [0.182, 0.388], n = 149 & $-$0.18 [$-$0.28, $-$0.07] \\
Structured & 0.848 [0.836, 0.862] & 0.0016 [0.0010, 0.0022] & 0.861 [0.820, 0.895], n = 309 & 0.405 [0.368, 0.440], n = 4,757 & $-$0.36 [$-$0.46, $-$0.27] \\
Verified & 0.836 [0.824, 0.847] & 0.0023 [0.0015, 0.0031] & 0.880 [0.853, 0.904], n = 320 & 0.694 [0.646, 0.737], n = 3,260 & $-$0.33 [$-$0.45, $-$0.22] \\
Unconstrained, Sonnet 5.5 & 0.900 [0.879, 0.920] & 0.0097 [0.0081, 0.0114] & 0.999 [0.997, 1.000], n = 107 & 0.168 [0.122, 0.217], n = 334 & +0.17 [$-$0.17, 0.41] \\
Explicit missingness, Sonnet 5.5 & 0.882 [0.865, 0.901] & 0.0001 [0.0000, 0.0005] & 0.999 [0.997, 1.000], n = 90 & 0.063 [0.020, 0.118], n = 206 & +0.75 [0.59, 0.90] \\
Structured, Sonnet 5.5 & 0.945 [0.934, 0.956] & 0.0014 [0.0007, 0.0021] & 1.000 [1.000, 1.000], n = 92 & 0.815 [0.765, 0.854], n = 2,482 & $-$0.53 [$-$0.66, $-$0.36] \\
Verified, Sonnet 5.5 & 0.952 [0.941, 0.962] & 0.0000 [0.0000, 0.0000] & 1.000 [1.000, 1.000], n = 92 & 0.999 [0.997, 1.000], n = 2,201 & $-$0.69 [$-$0.78, $-$0.53] \\
\bottomrule
\end{tabular}
\end{sidewaystable}

For the open-weight narrators, naming the absent modalities in the prompt was the single most effective intervention for acknowledgement and phantom evidence. Acknowledgement recall rose from 0.479 under the unconstrained regime to 0.881 under explicit missingness, and the phantom-evidence rate fell from 0.0206 to 0.0009, a factor of about twenty. The structured and verified regimes held acknowledgement near 0.84 and phantom evidence below 0.0025. A stricter variant of the phantom-evidence rate also includes the 35,091 claims that the extractor could not attribute to any modality, adding them to the denominator and counting the 2,341 of them that the pipeline scored as phantom evidence; it gave 0.039 under the unconstrained regime, 0.012 under explicit missingness, and below 0.0023 under the structured and verified regimes, so the ranking of regimes does not depend on the definition.

Attribution fidelity and value-of-information accuracy behaved differently. Judge-measured attribution fidelity was nearly constant across regimes, between 0.861 and 0.882 for the pooled open-weight narrators, and verification did not raise it: the difference between the verified and unconstrained regimes was 0.016 [$-$0.006, 0.038], and between the verified and explicit-missingness regimes $-$0.002 [$-$0.033, 0.027]. The verifier cannot correct what it does not detect, and in the judged sample the rule-based extractor identified 120 direction or ranking claims where the judge identified 8,615. Verification did reduce the fraction of reports containing at least one direction error, from 0.568 [0.509, 0.625] under the unconstrained regime to 0.416 [0.351, 0.483], but by shortening the reports rather than by correcting them: verified reports carried 5.6 direction claims on average against 8.8. Value-of-information accuracy was 0.387 and 0.268 under the first two regimes, on the small number of acquisition claims those regimes produced, 0.405 under the structured regime, whose template requires an acquisition statement for every absent modality, and 0.694 under verification. Confidence calibration was in the desired direction under every regime for the open-weight narrators, between $-$0.18 and $-$0.43.

The proprietary narrator differed in two respects. It acknowledged absent modalities in 0.900 of cases even without being told they were absent, so the explicit-missingness regime added nothing to acknowledgement, and its phantom-evidence rate under the unconstrained regime was half that of the pooled open-weight narrators. Its acquisition statements, however, were correct in only 0.168 of cases under the unconstrained regime and 0.063 under explicit missingness, and its stated confidence was positively correlated with model uncertainty under those two regimes, with Spearman's $\rho$ of +0.17 and +0.75, which is the wrong direction. Both failures were corrected by the structured template, which raised value-of-information accuracy to 0.815 and reversed the confidence correlation to $-$0.53, and by verification, which reached 0.999 on 2,201 acquisition claims with zero phantom evidence. Its judge-measured attribution fidelity was 0.999 or higher under every regime, with two contradicted claims among 4,228, so that direction errors were confined to the open-weight narrators.

Table~\ref{tab:3} separates the narrators under the structured regime, where every narrator produced acquisition claims in quantity. Acknowledgement recall was above 0.78 for every narrator and did not increase with scale among the open-weight models. Value-of-information accuracy did increase with scale, from 0.322 at 3 billion parameters to 0.463 at 14 billion and 0.815 for the proprietary model, as did judge-measured attribution fidelity, from 0.738 at 3 billion parameters to 0.880 at 7 billion, 0.920 at 14 billion, and 1.000 for the proprietary model. Under the unconstrained regime the pattern for phantom evidence was reversed: the 14-billion-parameter narrator cited absent modalities as evidence at a rate of 0.044 against 0.006 for the 3-billion-parameter narrator, because the larger model wrote more specific reports that named more modalities, while acknowledging absence in 0.689 of cases against 0.323.

\begin{sidewaystable}[htbp]
\centering
\small
\setlength{\tabcolsep}{4pt}
\caption{Narrative faithfulness by narrator under the structured regime, 3,200 reports per narrator, with the acknowledgement recall and phantom-evidence rate of the same narrator under the unconstrained regime in the last two columns. Attribution fidelity is judge-measured on a stratified random sample, with n the number of judged reports.}
\label{tab:3}
\begin{tabular}{lC{3.15cm}C{3.15cm}C{3.15cm}C{3.15cm}C{3.15cm}C{3.15cm}}
\toprule
Narrator & Acknowledgement recall & Phantom-evidence rate & Attribution fidelity, judge & Value-of-information accuracy & Acknowledgement, unconstrained & Phantom evidence, unconstrained \\
\midrule
Qwen2.5-3B & 0.890 [0.876, 0.903] & 0.0041 [0.0022, 0.0063] & 0.738 [0.625, 0.833], n = 106 & 0.322 [0.263, 0.387], n = 935 & 0.323 & 0.0058 \\
Qwen2.5-7B & 0.786 [0.765, 0.806] & 0.0009 [0.0002, 0.0019] & 0.880 [0.838, 0.917], n = 101 & 0.367 [0.310, 0.415], n = 1,487 & 0.424 & 0.0133 \\
Qwen2.5-14B & 0.868 [0.848, 0.887] & 0.0002 [0.0000, 0.0009] & 0.920 [0.889, 0.949], n = 102 & 0.463 [0.414, 0.501], n = 2,335 & 0.689 & 0.0437 \\
Claude Sonnet 5.5 & 0.945 [0.933, 0.956] & 0.0014 [0.0007, 0.0021] & 1.000 [1.000, 1.000], n = 92 & 0.815 [0.765, 0.858], n = 2,482 & 0.900 & 0.0097 \\
\bottomrule
\end{tabular}
\end{sidewaystable}

Table~\ref{tab:4} reports the claim-level hierarchical logistic regression on the 38,400 open-weight reports, with the unconstrained regime and the 3-billion-parameter narrator as reference levels. The odds of phantom evidence were lower by a factor of 10 to 23 under the three constrained regimes, rose by a factor of 2.37 for each additional absent modality, and rose by a factor of 2.03 for each unit of upstream phantom attribution mass, which supports the secondary hypothesis that phantom attribution in the model propagates into the narrative. The odds of a correct acquisition claim were 4.08 times higher under verification, 1.89 times higher for the 14-billion-parameter narrator, and were reduced by a factor of 0.43 per unit of phantom attribution mass. In the judged sample, the odds of a faithful direction claim were not raised by verification, with an odds ratio of 1.13 [0.98, 1.31], were raised slightly by explicit missingness, 1.19 [1.05, 1.36], and were 4.10 times higher for the largest open-weight narrator than for the smallest. They were multiplied by 0.94 for each additional absent modality and by 0.64 for each unit of upstream phantom attribution mass, so that phantom attribution in the model degrades the direction claims the narrator makes about the observed modalities as well as inviting it to cite the absent ones.

\begin{sidewaystable}[htbp]
\centering
\small
\setlength{\tabcolsep}{4pt}
\caption{Odds ratios with 95\% credible intervals from the claim-level Bayesian hierarchical logistic regression on the open-weight reports. Reference levels are the unconstrained regime and Qwen2.5-3B. Random intercepts for participant and run, except in the attribution-fidelity row, which is fitted on the 8,615 direction claims identified by the language-model judge in 1,233 sampled reports with a random intercept for participant. PAM is the phantom attribution mass of the upstream model under the background reference.}
\label{tab:4}
\begin{tabular}{p{2.6cm}C{2.75cm}C{2.75cm}C{2.75cm}C{2.75cm}C{2.75cm}C{2.75cm}C{2.75cm}}
\toprule
Outcome & Explicit & Structured & Verified & Qwen2.5-7B & Qwen2.5-14B & Per absent modality & Per unit PAM \\
\midrule
Phantom evidence & 0.043 [0.033, 0.056] & 0.069 [0.054, 0.089] & 0.102 [0.083, 0.126] & 1.29 [1.15, 1.44] & 3.53 [3.30, 3.78] & 2.37 [2.32, 2.41] & 2.03 [1.69, 2.43] \\
Attribution fidelity, judge & 1.19 [1.05, 1.36] & 0.94 [0.81, 1.08] & 1.13 [0.98, 1.31] & 1.42 [1.27, 1.58] & 4.10 [3.57, 4.71] & 0.94 [0.91, 0.97] & 0.64 [0.47, 0.87] \\
Value-of-information accuracy & 0.72 [0.49, 1.04] & 1.12 [1.06, 1.19] & 4.08 [3.77, 4.40] & 0.99 [0.91, 1.08] & 1.89 [1.77, 2.02] & 1.20 [1.17, 1.22] & 0.43 [0.36, 0.51] \\
\bottomrule
\end{tabular}
\end{sidewaystable}

\subsection{Value of Acquiring a Missing Modality}
\label{sec:5.3}

Restoring a masked modality changed the prediction by at least 0.10 in a substantial fraction of cases: amyloid PET in 26\% to 43\% of cases depending on the strategy, CSF in 22\% to 29\%, MRI in 20\% to 26\%, and APOE genotype in 3\% to 10\%. Table~\ref{tab:5} cross-tabulates stated against executed change at $\delta$ = 0.10 for the open-weight narrators pooled over regimes and for the proprietary narrator.

\begin{table}[htbp]
\centering
\footnotesize
\setlength{\tabcolsep}{4pt}
\caption{Acquisition claims against executed restoration at $\delta$ = 0.10, pooled over regimes. The overstatement rate is the fraction of cases with an executed change of at least $\delta$ for which the narrator stated that acquiring the modality would not change the assessment.}
\label{tab:5}
\begin{tabular}{llC{1.5cm}C{1.5cm}C{1.5cm}C{1.5cm}C{1.9cm}}
\toprule
Modality & Narrators & Stated: would change & Stated: would not change & Executed $\geq \delta$ & Executed $< \delta$ & Overstatement \\
\midrule
MRI & Open-weight & 1,380 & 803 & 532 & 1,653 & 0.102 \\
Amyloid PET & Open-weight & 2,436 & 493 & 1,000 & 1,929 & 0.028 \\
CSF & Open-weight & 1,439 & 506 & 514 & 1,431 & 0.023 \\
APOE genotype & Open-weight & 571 & 599 & 59 & 1,111 & 0.034 \\
MRI & Sonnet 5.5 & 310 & 1,034 & 217 & 1,127 & 0.184 \\
Amyloid PET & Sonnet 5.5 & 461 & 769 & 350 & 880 & 0.091 \\
CSF & Sonnet 5.5 & 315 & 798 & 214 & 899 & 0.051 \\
APOE genotype & Sonnet 5.5 & 91 & 1,059 & 31 & 1,119 & 0.258 \\
\bottomrule
\end{tabular}
\end{table}

The two families of narrator erred in opposite directions. The open-weight narrators stated that acquiring a modality would change the assessment in 71\% of their acquisition claims, whereas the restored modality in fact moved the prediction by at least $\delta$ in 26\% of cases, so their error was predominantly one of overstating the value of the missing test; their overstatement of sufficiency was rare, at 0.02 to 0.10. The proprietary narrator stated that acquisition would not change the assessment in 76\% of its claims, against an executed change rate of 17\%, and overstated sufficiency in 0.05 to 0.26 of cases where the restored modality in fact moved the prediction, most often for APOE genotype and MRI. At $\delta$ = 0.05 the proprietary narrator's overstatement rate rose to 0.28 to 0.70 and the open-weight narrators' to 0.06 to 0.19; at $\delta$ = 0.15 both fell, to at most 0.14 for MRI and 0.30 for genotype. The direction of the proprietary narrator's bias is the one reported for general-domain vision-language models \cite{ref11}; the open-weight narrators showed the opposite bias.

\subsection{Cost of Verification}
\label{sec:5.4}

For the open-weight narrators, 73\% of verified reports passed on the first attempt, 12\% needed one revision, 4\% two, and 10\% reached the three-round limit without passing every check; 90\% converged within two rounds. For the proprietary narrator 83\% passed at once and 99.3\% converged within two rounds. Verification shortened the open-weight reports, from a mean of 140 words under the unconstrained regime to 106 under verification, and reduced the fraction exceeding the 200-word limit from 1.6\% to 0.3\%. The proprietary narrator wrote about 188 words under every regime and exceeded the limit in 11\% to 28\% of reports. The claim extractor failed to parse 4.2\% of open-weight and 1.2\% of proprietary reports, and 6.0\% of extracted claims fell outside the taxonomy and were not verified. Clinical usefulness ratings were not collected.

\subsection{Validation of the Automatic Pipeline}
\label{sec:5.5}

Table~\ref{tab:6} reports agreement between the pipeline and each language-model judge and between the two judges. Acquisition statements and confidence levels were measured reliably: $\kappa$ between the pipeline and the judges was 0.63 and 0.77 for the stated direction of acquisition claims and the weighted $\kappa$ for confidence level was 0.77 and 0.78, with inter-judge $\kappa$ of 0.71 and 0.74. Acknowledgement agreement was 0.84 to 0.86 with precision 0.94 to 1.00 and recall 0.83 to 0.89; $\kappa$ was low, at 0.23 to 0.45, because most absent modalities were acknowledged and the statistic is penalised by the skewed base rate. Phantom evidence is a rare event, with 3 to 17 positive reports per 180 to 240 by either rater; raw agreement was 0.95 and $\kappa$ was 0.18 against the open-weight judge and 0.54 against the proprietary judge, which found more positives than the pipeline and whose positives the pipeline confirmed with precision 0.875. Direction errors were the weak point: both judges flagged many more direction errors than the pipeline, whose recall against them was 0.02 to 0.06, and the two judges agreed with each other only moderately, with $\kappa$ = 0.23. For this reason attribution fidelity in Tables~\ref{tab:2} to~\ref{tab:4} and Figure~\ref{fig:4} is measured by the language-model judge described in Section~\ref{sec:4.7} rather than by the pipeline, and the rule-based values are given in Supplementary Table S1.

\begin{table}[htbp]
\centering
\footnotesize
\setlength{\tabcolsep}{4pt}
\caption{Agreement between the automatic pipeline and two independent language-model judges, and between the judges. Precision and recall treat the judge as reference.}
\label{tab:6}
\begin{tabular}{p{0.22\linewidth}p{0.25\linewidth}p{0.25\linewidth}p{0.18\linewidth}}
\toprule
Quantity & Pipeline vs Qwen2.5-14B judge, n = 180 & Pipeline vs Sonnet 5.5 judge, n = 240 & Judge vs judge, n = 180 \\
\midrule
Acknowledgement of each absent modality & agreement 0.84, precision 1.00, recall 0.83, $\kappa$ 0.23 & agreement 0.86, precision 0.94, recall 0.89, $\kappa$ 0.45 & agreement 0.86, $\kappa$ 0.26 \\
Phantom evidence, report level & agreement 0.96, precision 0.14, recall 0.33, $\kappa$ 0.18 & agreement 0.95, precision 0.88, recall 0.41, $\kappa$ 0.54 & agreement 0.91, $\kappa$ 0.08 \\
Direction error, report level & agreement 0.65, precision 0.14, recall 0.02, $\kappa$ $-$0.04 & agreement 0.60, precision 0.75, recall 0.06, $\kappa$ 0.05 & agreement 0.61, $\kappa$ 0.23 \\
Stated direction of acquisition claim & agreement 0.84, $\kappa$ 0.63, n = 49 & agreement 0.89, $\kappa$ 0.77, n = 64 & agreement 0.86, $\kappa$ 0.71, n = 152 \\
Confidence level & weighted $\kappa$ 0.77, n = 22 & weighted $\kappa$ 0.78, n = 29 & weighted $\kappa$ 0.74, n = 158 \\
\bottomrule
\end{tabular}
\end{table}

The 50-report clinician-validated annotation points in the same direction as the judges and adds one failure mode that neither the pipeline nor the judges were asked to detect. Of the 92 absent-modality instances in the 38 reports with at least one absent modality, 87 were acknowledged, a recall of 0.946 against the pipeline's 0.886 on the same reports, with exact agreement on 87\% of reports. One report cited absent modalities as evidence, treating the missing MRI, PET, CSF, and genotype as reassuring negative findings; the pipeline did not flag it. Fifteen reports contained a statement about the value of acquiring a missing modality, seven asserting that it would not change the prediction and eight that it would help or hedging; the pipeline extracted an acquisition claim from one. Twenty-seven of the 50 reports contained at least one claim whose stated direction contradicted the attribution or misstated a fact about the patient, such as the patient's sex or amyloid status, against none flagged by the pipeline, which extracted a direction claim from only three of these prose reports. Four of the twelve reports with no absent modality, all from the 3- and 7-billion-parameter narrators, declared that observed modalities were unavailable, a false-absence error that the claim taxonomy does not contain. Usefulness ratings, confirmed by the physician, averaged 4.0 for the 14-billion-parameter narrator, 3.5 for the 7-billion, and 2.1 for the 3-billion, with 14 of 50 reports rated 1 or 2. The validated sheet is released with the code.

The judge used to re-measure attribution fidelity agreed with the validated annotation on the 24 calibration reports that fell in the judged batches. Agreement on whether a report contained a direction error was 0.92 with $\kappa$ = 0.82, nine positives by each, the rank correlation of the error counts was 0.84, and agreement on false absence, phantom evidence, and acknowledgement was complete. The two judges agreed closely with each other on the 100 reports that both judged: agreement on the presence of a direction error was 0.99 with $\kappa$ = 0.98, the rank correlation of the error counts was 0.98, attribution fidelity was 0.918 under both, agreement on acknowledgement and false absence was complete, and agreement on phantom evidence was 0.95 with $\kappa$ = 0.74. Attribution fidelity on the sampled open-weight reports was 0.881 for the reports judged by Claude Fable 5.1 and 0.863 for those judged by Claude Opus 5.5; since reports were assigned to batches at random, this difference is of the same order as the width of the cell-level intervals. The judge also flagged a form of phantom evidence that the pipeline scores as an acquisition claim: a statement that obtaining an absent modality would or would not change the assessment, justified by the attribution that the strategy assigned to its substitute, for example that acquiring amyloid PET is unlikely to matter because its contribution is small. Statements of this kind accounted for the phantom-evidence flags in 7.8\% of the sampled open-weight reports and 5.4\% of the proprietary reports under the verified regime, and they are the route by which phantom attribution lowers value-of-information accuracy in Table~\ref{tab:4}. Excluding them, the judge found phantom evidence in 17.5\% of the sampled unconstrained open-weight reports and in at most 0.6\% under each of the three constrained regimes, which confirms at report level the reduction that the pipeline measured at claim level.

Applied to all 38,400 open-weight reports, the false-absence detector flagged 141 reports, a rate of 0.37\%, but the rate was far from uniform. It was 3.0\% [2.4\%, 3.8\%] on the 2,400 reports for which no modality was absent, against 0.19\% [0.14\%, 0.23\%] when at least one was, so narrators invent a missing modality most often when there is none to name. It was highest under the explicit-missingness regime at 1.0\% [0.8\%, 1.2\%], whose prompt instructs the narrator to state which modalities are absent, next under the unconstrained regime at 0.46\% [0.32\%, 0.60\%], and zero under the structured and verified regimes, whose template supplies a field that can be filled with none. It fell with narrator scale, from 0.69\% for the 3-billion-parameter narrator to 0.30\% and 0.12\%, and under the explicit regime reached 1.8\% [1.3\%, 2.3\%] for the smallest narrator. The modalities most often falsely declared absent were APOE genotype, CSF, and demographics. Table~\ref{tab:7} gives the rates by narrator and regime.

\begin{table}[htbp]
\centering
\footnotesize
\setlength{\tabcolsep}{4pt}
\caption{False-absence rate by narrator and regime over 3,200 reports per cell, with 95\% bootstrap intervals and the number of flagged reports. A false absence is a statement that an observed modality was unavailable.}
\label{tab:7}
\begin{tabular}{lC{3.2cm}C{3.2cm}C{3.2cm}C{3.2cm}}
\toprule
Narrator & Unconstrained & Explicit missingness & Structured & Verified \\
\midrule
Qwen2.5-3B & 0.97\% [0.66\%, 1.31\%], n = 31 & 1.78\% [1.34\%, 2.25\%], n = 57 & 0, n = 0 & 0, n = 0 \\
Qwen2.5-7B & 0.09\% [0.00\%, 0.22\%], n = 3 & 1.09\% [0.75\%, 1.47\%], n = 35 & 0, n = 0 & 0, n = 0 \\
Qwen2.5-14B & 0.31\% [0.16\%, 0.50\%], n = 10 & 0.16\% [0.03\%, 0.31\%], n = 5 & 0, n = 0 & 0, n = 0 \\
\bottomrule
\end{tabular}
\end{table}

\subsection{Natural Missingness}
\label{sec:5.6}

Table~\ref{tab:8} compares the 60 naturally incomplete participants, 900 reports per regime from the open-weight narrators, with the complete-data reports restricted to the same seven missingness patterns, 4,200 per regime. The two agreed closely. Acknowledgement recall under the unconstrained regime was 0.443 on natural against 0.496 on designed missingness, and 0.86 to 0.87 under the three constrained regimes on both; the phantom-evidence rate was 0.0097 against 0.0131 under the unconstrained regime and at most 0.0004 against at most 0.0013 under the others; confidence calibration also overlapped within its intervals. Attribution fidelity is not compared, because the judged sample was drawn from the complete-data reports only. The designed protocol therefore reproduced the narrator behaviour observed under the missingness that ADNI actually exhibits.

\begin{table}[htbp]
\centering
\footnotesize
\setlength{\tabcolsep}{4pt}
\caption{Natural against designed missingness for the open-weight narrators, restricted to the seven patterns present in the natural sub-sample: PET with CSF in 26 participants, PET in 23, CSF in 6, and four rarer patterns in 5. Attribution fidelity is not reported for this comparison because the judged sample contains complete-data reports only.}
\label{tab:8}
\begin{tabular}{p{2.3cm}C{2.9cm}C{2.9cm}C{2.9cm}C{2.9cm}}
\toprule
Regime & Acknowledgement, natural & Acknowledgement, designed & Phantom evidence, natural & Phantom evidence, designed \\
\midrule
Unconstrained & 0.443 [0.395, 0.500] & 0.496 [0.475, 0.515] & 0.0097 [0.0056, 0.0136] & 0.0131 [0.0103, 0.0163] \\
Explicit missingness & 0.860 [0.821, 0.894] & 0.883 [0.865, 0.900] & 0.0004 [0.0000, 0.0010] & 0.0008 [0.0003, 0.0015] \\
Structured & 0.866 [0.829, 0.899] & 0.866 [0.854, 0.878] & 0.0000 [0.0000, 0.0000] & 0.0013 [0.0005, 0.0023] \\
Verified & 0.864 [0.832, 0.894] & 0.856 [0.841, 0.870] & 0.0000 [0.0000, 0.0000] & 0.0010 [0.0004, 0.0018] \\
\bottomrule
\end{tabular}
\end{table}

\subsection{Sensitivity Analyses}
\label{sec:5.7}

The reference-value ablation is reported in Table~\ref{tab:1} and discussed in Section~\ref{sec:5.1}. The threshold ablation is reported in Section~\ref{sec:5.3}. Under the rephrased prompt, on 665 reports per narrator, the regime effect was preserved for every narrator: acknowledgement recall from the unconstrained to the structured regime moved from 0.123 to 0.896 for Qwen2.5-3B, from 0.232 to 0.606 for Qwen2.5-7B, from 0.616 to 0.813 for Qwen2.5-14B, and from 0.908 to 0.959 for the proprietary narrator, and the phantom-evidence rate fell to at most 0.007 under the structured regime for every narrator. Across five sampling seeds at temperature 0.7 under the structured regime, on 328 reports per seed, the between-seed standard deviation was at most 0.035 for acknowledgement recall, 0.004 for the phantom-evidence rate, and 0.057 for value-of-information accuracy; it was 0.11 to 0.16 for the rule-based attribution fidelity, which rests on few claims per seed.

\section{Discussion}
\label{sec:6}

The central claim of this work is that an explanation of a multimodal prediction can be corrupted by missing evidence at two points that are usually conflated, and that the two require different remedies. The results bear this out, and they also correct the expectations with which the study was designed.

The first correction concerns the model. We expected architectures that drop an absent modality's token, or replace it with a learned placeholder, to assign little attribution to the absent modality, and architectures that impute to assign much. Instead every strategy assigned between 12\% and 23\% of its attribution mass to absent modalities, and the ordering was not the expected one: the indicator strategy was lowest, multiple imputation highest, and the placeholder and dropout strategies were indistinguishable from each other and, under the background reference, from the indicator. The reason is that the definition in Section~\ref{sec:3.2} measures the attribution assigned to the player that represents an absent modality, and that player's value is the effect of the modality's absence on the prediction relative to the reference. For an imputing strategy that effect is partly the information content of the imputed values, which is fabricated. For a dropout or placeholder strategy it is the shift in the prediction that follows from the model knowing that the modality is missing, which in a cohort where PET and CSF are missing not at random is informative rather than fabricated \cite{ref22}. Phantom attribution mass as defined here captures both, and the present data cannot separate them. What the data do show is that the explanation handed to a narrator or a clinician assigns substantial credit to modalities that were not measured under every strategy in common use, that the credit grows with each additional absent modality, and that the largest share goes to amyloid PET, the modality most often missing in practice. The reference ablation adds a second lesson: under the pre-specified mean reference, mean imputation appears to have no phantom attribution at all, purely because the substitute and the reference coincide. A report of phantom attribution mass is uninterpretable without the reference, and we recommend the background-sample reference as the default.

The second correction concerns the narrator. We expected explicit prompting to be necessary but not sufficient, and the first half held: for the open-weight narrators, naming the absent modalities in the prompt raised acknowledgement from 48\% to 88\% and cut phantom evidence twenty-fold, and no amount of additional structure or verification improved acknowledgement further. The second half held only for acquisition claims. Once told that a modality is absent, the open-weight narrators rarely cited it as evidence, but their statements about whether acquiring it would change the assessment were correct in 27\% to 41\% of cases, and the proprietary narrator's were correct in 6\% to 17\% under the two prose regimes. These statements have no basis in the structured input, because nothing in it says what the model would do with the missing modality, and the narrators filled the gap in opposite directions: the open-weight models asserted that the missing test would change the assessment far more often than it would, and the proprietary model asserted that the available evidence sufficed, which is the bias reported for general-domain vision-language models \cite{ref11}. The structured template improved the proprietary narrator substantially and the open-weight narrators little; verification against executed restorations improved both, to 69\% and 99.9\%, because it supplies the one thing the narrator lacks. The practical conclusion is that acquisition claims should be generated from executed restorations rather than inferred by the narrator, which is feasible whenever the model can be re-run on a hypothetical value, or should be omitted.

The mixed-effects analysis links the two failure points. Each unit of phantom attribution mass in the upstream model doubled the odds that the narrative cited an absent modality as evidence and more than halved the odds that an acquisition claim was correct, and in the judged sample it reduced by a third the odds that a direction claim about an observed modality was faithful. A narrator handed an attribution table in which an unmeasured modality carries weight will reproduce that weight in prose, and a prompt that says the modality is absent does not fully override a table that says it mattered. Reducing phantom attribution at the model is therefore a prerequisite for faithful narration and not a substitute for it.

The contrast between designed and natural missingness supports the protocol. On the 60 participants whose modalities were genuinely unmeasured, narrator behaviour under every regime matched that on the complete-data participants with the same modalities masked, within the intervals. The masking protocol can be applied to any cohort with a complete-data subset and yields ground truth that a naturally incomplete cohort cannot, without, on this evidence, misrepresenting what happens on naturally incomplete patients.

The cost of verification was modest. Nine in ten open-weight reports and all but 0.7\% of proprietary reports converged within two rounds, verified reports were shorter and more often within the length limit, and the parse failure rate was 4\% or less. Whether verification changes the clinical usefulness of a report is not answered here, because the clinician ratings in the protocol were not collected.

More generally, the study illustrates a division of labour between interventional and human evaluation that we believe is the appropriate one for language-model explanations in medicine. The metrics are defined by what the model does under intervention, so that they can be computed at scale and cannot be inflated by fluency \cite{ref10}. Independent judgement validates the machinery that computes them. Here that judgement came from two language-model judges and from a clinician-validated annotation of 50 reports, and Section~\ref{sec:5.5} shows where the pipeline was adequate, for acquisition statements, confidence, and acknowledgement, and where it was not, for direction errors, which it detected in none of the 50 reports in which the validated annotation found them in 27. Replacing the pipeline with a calibrated judge for that one metric changed a substantive conclusion: verification, which by the pipeline's measure made direction claims almost perfectly faithful, did not change their fidelity, because the verifier checked only the few direction claims that the extractor found. Interventional metrics are only as reliable as the claim extraction beneath them, and the extraction of direction claims from prose, where direction is carried by headings and list framing rather than by single sentences, is where a rule-based pipeline is weakest. The annotation also surfaced false absence, which the taxonomy did not anticipate and which the explicit-missingness prompt makes worse: told to name what is missing, the smallest narrators name something even when nothing is. A template with a field that can be left empty removed the error entirely.

\section{Limitations}
\label{sec:7}

The interventional ground truth is defined with respect to the prognostic model, not with respect to the patient. An acquisition claim is scored as correct when restoring the modality changes the model's prediction, which is the appropriate standard for faithfulness but is silent on whether the model's response to the restored modality is itself clinically correct. Predictive performance is reported beside every faithfulness metric for this reason, and the protocol should be read as evaluating the explanation of a given model rather than the model.

Phantom attribution mass conflates attribution to fabricated values with attribution to the fact of missingness, as Section~\ref{sec:6} sets out. Separating the two would require an attribution method that conditions on the observed modalities rather than substituting for the absent ones, or a cohort in which missingness is known to be uninformative, and neither was available here. The definition is also tied to the reference value, and we have shown that one common choice makes the quantity vanish by construction for one common strategy.

The pipeline's detection of direction errors is poor, and attribution fidelity is therefore measured by a language-model judge on a sample of about 100 reports per narrator and regime rather than by the pipeline on the full corpus. The judge is itself a language model. Its calibration rests on 24 reports whose reference annotation was drafted by Claude Fable 5.1, one of the two judges, and confirmed by a physician, so the agreement on those reports is not independent of the judge, and the close agreement between the two judges shows that the measurement is reproducible across judges rather than that it is correct. The judge assesses the direction of claims and not their ranking, so judge-measured and rule-based fidelity are not the same quantity, and the judged sample does not include the naturally incomplete participants.

The protocol called for two human annotators on 200 reports and two physicians rating 50. What exists is validation against language-model judges, one of which is also a narrator, and a 50-report annotation drafted by a language model and confirmed by one physician. A single reviewer confirming a draft is weaker evidence than two independent annotators, since agreement with a draft is easier than agreement from scratch, and the usefulness ratings reflect one physician's judgement of 50 reports from three narrators under two regimes. The direction-error rate of 54\% in those 50 reports is an estimate from a small sample drawn mostly from one regime; on the stratified judged sample the corresponding rate for the explicit-missingness regime was 0.468 [0.405, 0.526].

The narration study rests on 40 participants from the complete-data cohort and 60 from the naturally incomplete cohort, with every participant narrated many times. Reports from the same participant are correlated, which the clustered bootstrap and the random intercepts address, but the number of distinct clinical situations is small, and twelve converters among the 40 limits what can be said about converters specifically.

The complete-data cohort is a selected subset of ADNI. Participants with every modality observed at baseline are those who consented to PET and lumbar puncture within the study phases that offered them, and they differ in age, enrolment phase, and disease stage from the cohort as a whole. The agreement between designed and natural missingness in Table~\ref{tab:8} is reassuring but rests on 60 participants and seven patterns.

The inclusion of a proprietary narrator departs from the protocol, which restricted narrators to open-weight models on data-governance grounds. Its results are reported separately and should be read as an indication of how a larger model behaves rather than as part of the pre-specified comparison. The open-weight narrators are all from one model family, and the narration regimes are four points on a continuum; fine-tuning a narrator on verified reports, or generating acquisition claims directly from executed restorations, were not evaluated.

Finally, the outcome is a clinical diagnosis of AD dementia at 36 months, which depends on clinician judgement and on the follow-up schedule. The 715 participants lost to follow-up before 36 months are excluded, and the sensitivity analysis that retains them as non-converters was not run.

\section{Conclusions}
\label{sec:8}

Multimodal prognostic models will be used on patients for whom some modalities are missing, and their explanations will be read by clinicians who need to know what the model saw and what it did not. This paper separates two ways in which a missing modality can corrupt such an explanation, phantom attribution in the model and unfaithful narration by the language model, defines both as quantities with interventional ground truth, and measures them on ADNI across five missing-data strategies, four narration regimes, and 51,200 reports. No strategy in common use was free of phantom attribution, which rose with each absent modality and fell most heavily on amyloid PET; a missingness indicator was the least affected. Naming the absent modalities in the prompt was sufficient to make narrators acknowledge them and stop citing them, but no prompt made narrators' statements about which test to order next reliable, because the information those statements require is not in the attribution table; verifying each claim against the model's behaviour under restoration was. Phantom attribution in the model propagated into phantom evidence in the narrative, and a clinician-validated annotation showed that direction errors are far more common than the rule-based verifier detects and that narrators told to name absent modalities sometimes invent them. Measured by a calibrated judge, the fidelity of direction claims depended on the scale of the narrator and on the phantom attribution of the model, and not on the prompting regime or on verification. The masking protocol reproduced narrator behaviour on naturally incomplete patients. We propose that phantom attribution mass, with its reference stated, and acquisition-claim accuracy against executed restoration be reported wherever explanations of multimodal clinical predictions are offered under missing data, and we release the protocol and code so that they can be.

\section*{Declarations}

\paragraph*{Author Contributions} Conceptualization, F.K.; methodology, F.K.; software, F.K.; validation, F.K.; formal analysis, F.K.; data curation, F.K.; writing, original draft preparation, F.K.; writing, review and editing, F.K. The author has read and agreed to the published version of the manuscript.

\paragraph*{Funding} This research received no external funding.

\paragraph*{Institutional Review Board Statement} This study is a secondary analysis of de-identified data obtained under the ADNI data use agreement. ADNI study procedures were approved by the institutional review boards of all participating institutions.

\paragraph*{Informed Consent Statement} Written informed consent was obtained from all ADNI participants at each participating site.

\paragraph*{Data Availability Statement} ADNI data are available to qualified researchers at adni.loni.usc.edu upon application. Code for the masking protocol, models, claim extraction and verification, and analysis are available from the corresponding author on request and will be released publicly upon publication. Derived data that cannot be redistributed under the ADNI agreement will be described by the scripts that generate them.

\paragraph*{Acknowledgments} Data collection and sharing for this project were funded by the Alzheimer's Disease Neuroimaging Initiative, whose current funder is the National Institute on Aging through National Institutes of Health grant U19AG024904, with the Northern California Institute for Research and Education as grantee organisation. Past funding came from the National Institute of Biomedical Imaging and Bioengineering, the Canadian Institutes of Health Research, and private sector contributions facilitated by the Foundation for the National Institutes of Health from AbbVie, Alzheimer's Association, Alzheimer's Drug Discovery Foundation, Araclon Biotech, BioClinica, Inc., Biogen, Bristol-Myers Squibb Company, CereSpir, Inc., Cogstate, Eisai Inc., Elan Pharmaceuticals, Inc., Eli Lilly and Company, EuroImmun, F. Hoffmann-La Roche Ltd and its affiliated company Genentech, Inc., Fujirebio, GE Healthcare, IXICO Ltd., Janssen Alzheimer Immunotherapy Research \& Development, LLC., Johnson \& Johnson Pharmaceutical Research \& Development LLC., Lumosity, Lundbeck, Merck \& Co., Inc., Meso Scale Diagnostics, LLC., NeuroRx Research, Neurotrack Technologies, Novartis Pharmaceuticals Corporation, Pfizer Inc., Piramal Imaging, Servier, Takeda Pharmaceutical Company, and Transition Therapeutics. ADNI data are disseminated by the Laboratory for Neuro Imaging at the University of Southern California.

\paragraph*{Use of Generative AI} Claude (Anthropic) was used to edit the text and to assist with coding. The author reviewed all output and is responsible for the entire content of the paper.

\paragraph*{Conflicts of Interest} The author declares no conflict of interest.

\end{document}